\documentclass{article}

\PassOptionsToPackage{numbers,compress}{natbib}
\usepackage[preprint]{neurips_2026} 

\usepackage[utf8]{inputenc}
\usepackage[T1]{fontenc}
\usepackage{hyperref}
\usepackage{url}
\usepackage{booktabs}
\usepackage{amsfonts}
\usepackage{amsmath}
\usepackage{amssymb}
\usepackage{nicefrac}
\usepackage{microtype}
\usepackage{xcolor}
\usepackage{graphicx}
\usepackage{multirow}
\usepackage{wrapfig}
\usepackage{xspace}
\usepackage{enumitem}
\usepackage{algorithm}
\usepackage{algorithmic}
\usepackage{natbib}

\title{Elastic-dLLM: Position Preserving Context Compression and Augmentation of Diffusion LLMs}

\author{
  Junyi Wu$^{1,2,*,\dagger}$~,\enspace
  Tianchen Zhao$^{1,*}$~,\enspace
  Shaoqiu Zhang$^{2}$,\enspace
  Linfeng Zhang$^{2}$,\enspace
  Guohao Dai$^{2,3}$,\enspace \\
  \textbf{Yu Wang}$^{1,\ddagger}$~\enspace \\
  \textsuperscript{1}Tsinghua University\enspace
  \textsuperscript{2}Shanghai Jiao Tong University\enspace
  \textsuperscript{3}Infinigence AI\enspace
  \vspace{-4mm}
}

\begin{document}

\maketitle
\begingroup
\renewcommand{\thefootnote}{\fnsymbol{footnote}}
\footnotetext[1]{\scriptsize Equal contribution.}
\footnotetext[2]{\scriptsize Work done during an internship at Tsinghua University.}
\footnotetext[3]{\scriptsize Corresponding author.}
\endgroup

\begin{abstract}

Unlike autoregressive models, which generate one token at a time, dLLMs denoise a chunk of \texttt{[MASK]} tokens jointly and sample one or more tokens per step; despite enabling parallel decoding, this process incurs substantial computational cost due to the large chunk size of masked tokens. We observe that much of this cost is spent on repeatedly processing the preceding context and many \texttt{[MASK]} tokens with the same feature representations, indicating considerable computational redundancy. In this work, we revisit dLLM's redundancy from the perspective of \texttt{[MASK]} tokens. Through systematic analysis, we verify the redundancy of \texttt{[MASK]} tokens while revealing their critical role in providing structural information. Guided by these findings, we propose position-preserving \texttt{[MASK]} token compression and terminal-aware augmentation. By compressing redundant \texttt{[MASK]} computation, this approach accelerates decoding and further provides a natural extension toward context-folding-like long-context scaling under limited input-length constraints for full-sequence dLLMs such as LLaDA-8B-Instruct and LLaDA-1.5. Moreover, for block dLLMs such as LLaDA2.0-mini, it augments the context with a protected terminal \texttt{[MASK]} token to enhance generation quality with negligible overhead.

\end{abstract}

\begin{figure}[H]
    \centering
    \includegraphics[width=0.95\linewidth]{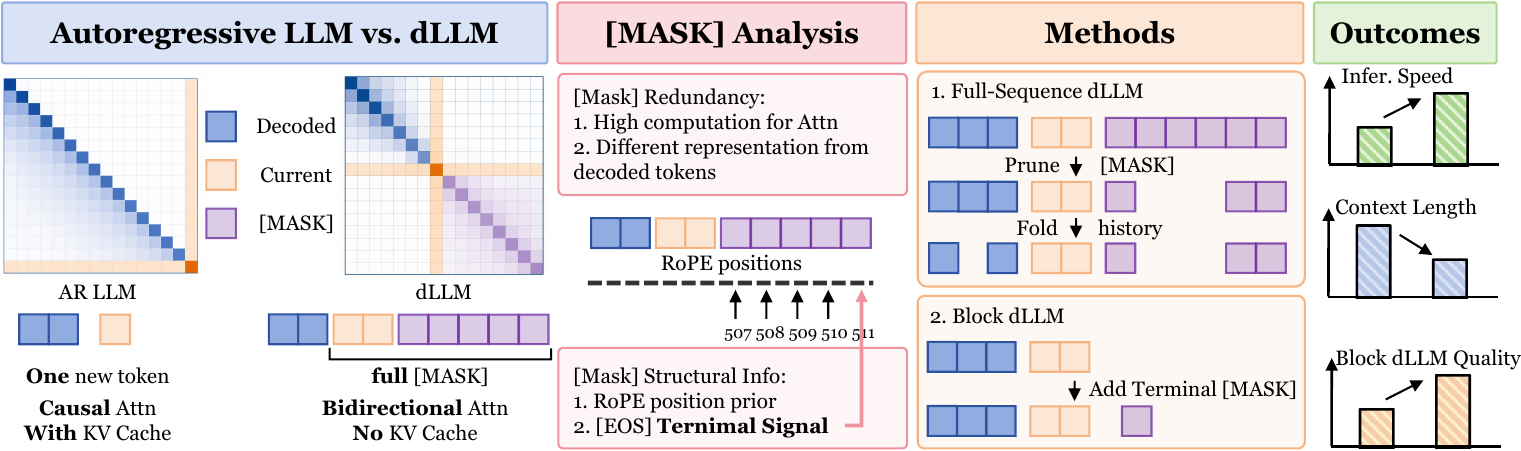}
    \caption{Compared with autoregressive decoding, dLLMs repeatedly compute bidirectional attention over a preallocated \texttt{[MASK]} context. Elastic-dLLM compacts this context to accelerate full-sequence inference, enable test-time context scaling, and improve block dLLMs with one protected terminal token.}
    \label{fig:teaser}
    \vspace{-10pt}
\end{figure}

\section{Introduction}
\label{sec:intro}

Diffusion language models (dLLMs)~\cite{llada,llada1.5,llada2} have emerged as a promising alternative to autoregressive language models. Instead of left-to-right factorization, they generate text through iterative denoising. This difference changes the structure of inference, as illustrated in Fig.~\ref{fig:teaser}. Autoregressive models use causal attention over a decoded prefix, which makes key-value caching a natural way to reuse computation across steps. In contrast, dLLMs use bidirectional attention over a preallocated generation context that contains decoded tokens and \texttt{[MASK]} positions. A \texttt{[MASK]} token marks a position where the content has not yet been generated and remains undenoised. Similar to \texttt{[EOS]}, it acts as special token in the vocabulary with its own unique feature representation. Depending on how this context is organized during inference, existing dLLMs mainly follow two schemes. \emph{Full-sequence dLLMs}~\citep{llada,llada1.5} preallocate all \texttt{[MASK]} positions up to the planned generation length and denoise them together. The decoded history, the current generation region, and the remaining \texttt{[MASK]} positions are processed in the same active context. \emph{Block dLLMs}~\citep{llada2} use only the current block of \texttt{[MASK]} positions, denoise that block, and then append the next block.

Although dLLMs enable parallel decoding, they incur additional computation compared with autoregressive LLMs, since they process a full chunk of \texttt{[MASK]} tokens to generate only one or a few tokens. This cost is substantially higher in the full-sequence scheme, where \texttt{[MASK]} tokens become a dominant source of inference overhead. In particular, a substantial amount of computation is spent on a large number of \texttt{[MASK]} tokens that share the same feature representation, suggesting significant redundancy. This raises a key question: \textbf{What role do [MASK] tokens play in dLLM decoding, and how can they be compacted without sacrificing generation quality?}

To answer that, we conduct extensive analysis on the role of \texttt{[MASK]} tokens and reveal that they are both \textbf{redundant} and \textbf{structurally informative}, as illustrated in Fig.~\ref{fig:teaser}. 
Based on these observations, we propose Elastic-dLLM, a training-free framework that adjusts the dLLM context to reduce redundancy while exploiting the structural information encoded in \texttt{[MASK]} tokens. Fig.~\ref{fig:teaser} gives an overview of the Elastic-dLLM framework. For context compression, we design a position-preserving compression technique that maintains the structural information carried by \texttt{[MASK]} tokens while eliminating most of the redundant computation. This compression scheme can be naturally extended to a context-folding-like~\cite{sun2025scaling,wu2025resum,kang2025acon} context scaling method for full-sequence dLLMs with a fixed maximum input length, thereby further enhancing performance. 
For context augmentation, we consider block dLLMs, where decoding is performed over a single active block. Although efficient, this scheme weakens the model's awareness of the ``terminal signal'', which indicates when the generation has to stop. It can lead to failure cases in long-context generation, where the model over-generates and repeats the final segment.
Motivated by the structural information encoded in \texttt{[MASK]} tokens, we introduce a protected, non-decoded terminal \texttt{[MASK]} token whose RoPE embedding corresponds to the maximum generation position. This lightweight augmentation restores the missing terminal signal in block decoding and improves generation quality with negligible overhead.

We summarize our key contributions as follows:
\begin{enumerate}
    \item \textbf{We analyze the role of \texttt{[MASK]} tokens in dLLM decoding.}
    Through extensive analysis, we show that \texttt{[MASK]} tokens are not merely redundant placeholders, but also encode structural information that is essential for decoding.
    
    \item \textbf{Position-Preserving Context Compression and Scaling.}
    Elastic-dLLM introduces a position-preserving compression strategy that preserves structural information while reducing cost. Beyond compression, the same mechanism can be naturally extended into a context-folding-like scaling method to further improve generation quality.

    \item \textbf{Terminal-aware Context Augmentation.} Elastic-dLLM compensates for the missing terminal signal by introducing an additional \texttt{[MASK]} token at the terminal RoPE position, effectively mitigating repetition of the final segment.
\end{enumerate}

\section{Related Work}
\label{sec:related}

\paragraph{Diffusion language models.}
Discrete diffusion replaces left-to-right generation with iterative denoising. SEDD~\citep{sedd} develops score-entropy training for discrete diffusion, and masked diffusion language models~\citep{mdlm} show that absorbing-state masking can support flexible generation while approaching autoregressive perplexity. Recent dLLMs such as LLaDA~\citep{llada,llada1.5} and Dream~\cite{dream} scale masked denoising to modern language-model settings. Block diffusion further denoises one block at a time, improving variable-length generation and making part of the decoding process closer to autoregressive execution~\citep{blockdiffusion}. These models change the inference problem: generation is no longer a sequence of single next-token updates, but repeated denoising over a planned context that may contain many future \texttt{[MASK]} tokens. This raises a practical question that is less central in autoregressive decoding: which parts of the masked future context are actually necessary during inference?

\paragraph{Efficient inference for dLLMs.}
Existing efficient-inference methods for dLLMs exploit different forms of redundancy in iterative denoising. Cache-based methods reuse stable intermediate
representations or KV states across nearby denoising steps, including delayed
caching, prompt--response caching, guided reuse, dual adaptive caching, and
attention-aware refresh~\citep{dkvcache,dllmcache,flashdlm,elasticcache,d2cache,quantcache,quantdllm,flashblock}.
Hybrid cache-scheduling methods such as Fast-dLLM~\citep{fastdllm} combine
approximate block-wise KV reuse with confidence-aware parallel decoding.
Scheduling and pruning methods adaptively decide when tokens should be
committed, refreshed, or skipped~\citep{esdllm}, while sparse-attention methods
reduce per-step cost using stable or local attention patterns~\citep{sparsed}.
These methods mainly target state recomputation, denoising schedules, or dense
attention. Despite their effectiveness, they typically rely on the empirical
stability of hidden states, attention maps, or token confidence across adjacent
denoising steps. As a result, cache validity, refresh granularity, and decoding
schedule design become key factors in balancing inference speed, memory overhead,
and generation quality.
Less explored is the denoising layout itself: full-sequence dLLMs repeatedly process a long context filled with future \texttt{[MASK]} positions, so reducing active context length could be another source of efficiency if the shortened layout preserves the structural information carried by those positions.

\paragraph{Long-context inference and token compression.}
Long-context autoregressive inference has motivated many token and KV-cache compression methods. StreamingLLM~\citep{streamingllm} preserves attention-sink tokens for stable streaming generation; H2O~\citep{h2o} and Scissorhands~\citep{scissorhands} retain heavy-hitter or persistently important tokens; SnapKV and Quest~\citep{snapkv,quest} select cache tokens or pages according to prompt or query attention patterns; and PyramidKV, KIVI, and InfLLM~\citep{pyramidkv,kivi,infllm} explore layer-wise budgeting, cache quantization, or external memory for long sequences. These methods assume causal decoding, where past keys and values are computed once and then retained, pruned, or compressed for future queries. The compressed objects are therefore usually decoded prefix states, and their utility is measured by how well they support later next-token queries without recomputing the full history. dLLMs pose a related but different question because of the planned \texttt{[MASK]} tokens and bidirectional attention, leaving open how token compression should be defined for this denoising layout.

\section{The Role of \texttt{[MASK]} Tokens in dLLM Decoding}
\label{sec:observation}

In this section, we present an extensive analysis of the role of \texttt{[MASK]} tokens, showing that they exhibit high redundancy while preserving essential structural information. As seen in Fig.~\ref{fig:overview}, future \texttt{[MASK]} tokens actively participate in attention (Section~\ref{sec:mask_token_cost}), yet many of them carry similar non-semantic representations and are therefore compressible (Section~\ref{sec:decoded_mask_space}). At the same time, their original RoPE coordinates influence what content is decoded at a position (Section~\ref{sec:rope_observation}), and the terminal region provides boundary information (Section~\ref{sec:eos_observation}). Therefore, compacting dLLM context cannot be ordinary token pruning; it must preserve the positional skeleton of the generation context.

\begin{figure}[t]
    \centering
    \begin{minipage}[t]{0.48\linewidth}
        \centering
        \includegraphics[width=0.6\linewidth]{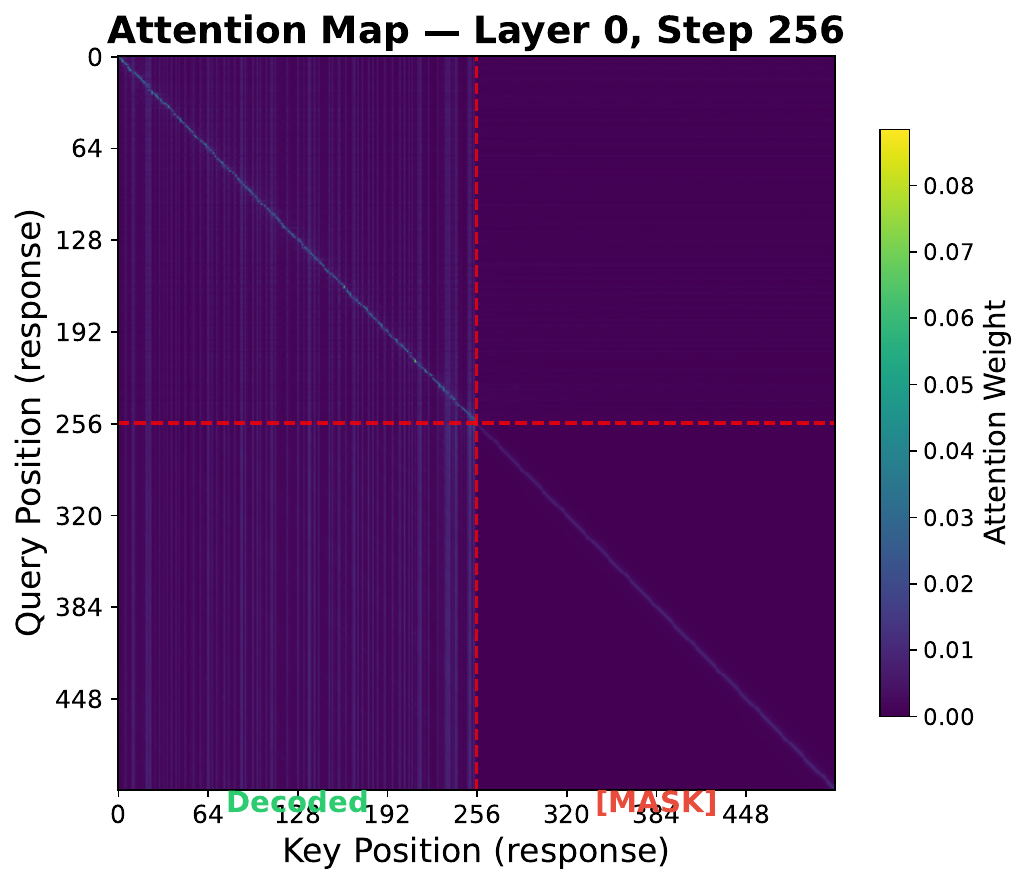}
        \caption{\textbf{\texttt{[MASK]} Attention Redundancy.} Attention maps during denoising show sparse and regular patterns: \texttt{[MASK]} tokens mainly aggregate locally among themselves, while their interactions with decoded tokens are relatively weak.}
        \label{fig:mask_attention}
    \end{minipage}
    \hfill
    \begin{minipage}[t]{0.48\linewidth}
        \centering
        \includegraphics[width=\linewidth]{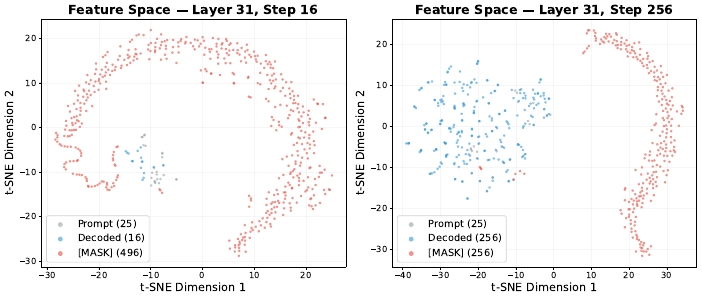}
        \caption{\textbf{Representation Separation.} t-SNE visualization of hidden states from decoded tokens and \texttt{[MASK]} tokens. The two groups form separated clusters, indicating that \texttt{[MASK]} tokens occupy a distinct feature space from decoded semantic tokens.}
        \label{fig:tsne_mask_space}
    \end{minipage}
\end{figure}

\subsection{\texttt{[MASK]} tokens have redundancy}
\label{sec:mask_redundancy}
\label{sec:mask_token_cost}
\label{sec:decoded_mask_space}

\noindent\textbf{Attention Map of \texttt{[MASK]} Tokens.}\quad
We visualize the attention patterns in the middle stage of generation in Fig.~\ref{fig:mask_attention}. As shown, the \texttt{[MASK]} tokens exhibit highly sparse and regular attention patterns. Specifically, the attention map reveals a characteristic structure: \texttt{[MASK]} tokens mainly aggregate information locally among themselves (the lower right); their interactions with decoded tokens are relatively weak (the upper right), \textbf{validating their high redundancy.}

\noindent\textbf{Feature-space separation between \texttt{[MASK]} and decoded tokens:}\quad
Motivated by the distinctive attention patterns of \texttt{[MASK]} tokens, we further investigate the relationship between \texttt{[MASK]} tokens and decoded tokens in the feature space. To this end, we perform a t-SNE visualization of their hidden features, as shown in Fig.~\ref{fig:tsne_mask_space}. The features naturally form two clearly separated clusters, indicating that \texttt{[MASK]} tokens and decoded tokens exhibit distinct feature-space properties. This suggests that \textbf{\texttt{[MASK]} tokens play a different role from normal decoded tokens during dLLM decoding.}

\begin{figure}[t]
    \centering
    \begin{minipage}[t]{0.48\linewidth}
        \centering
        \includegraphics[width=\linewidth]{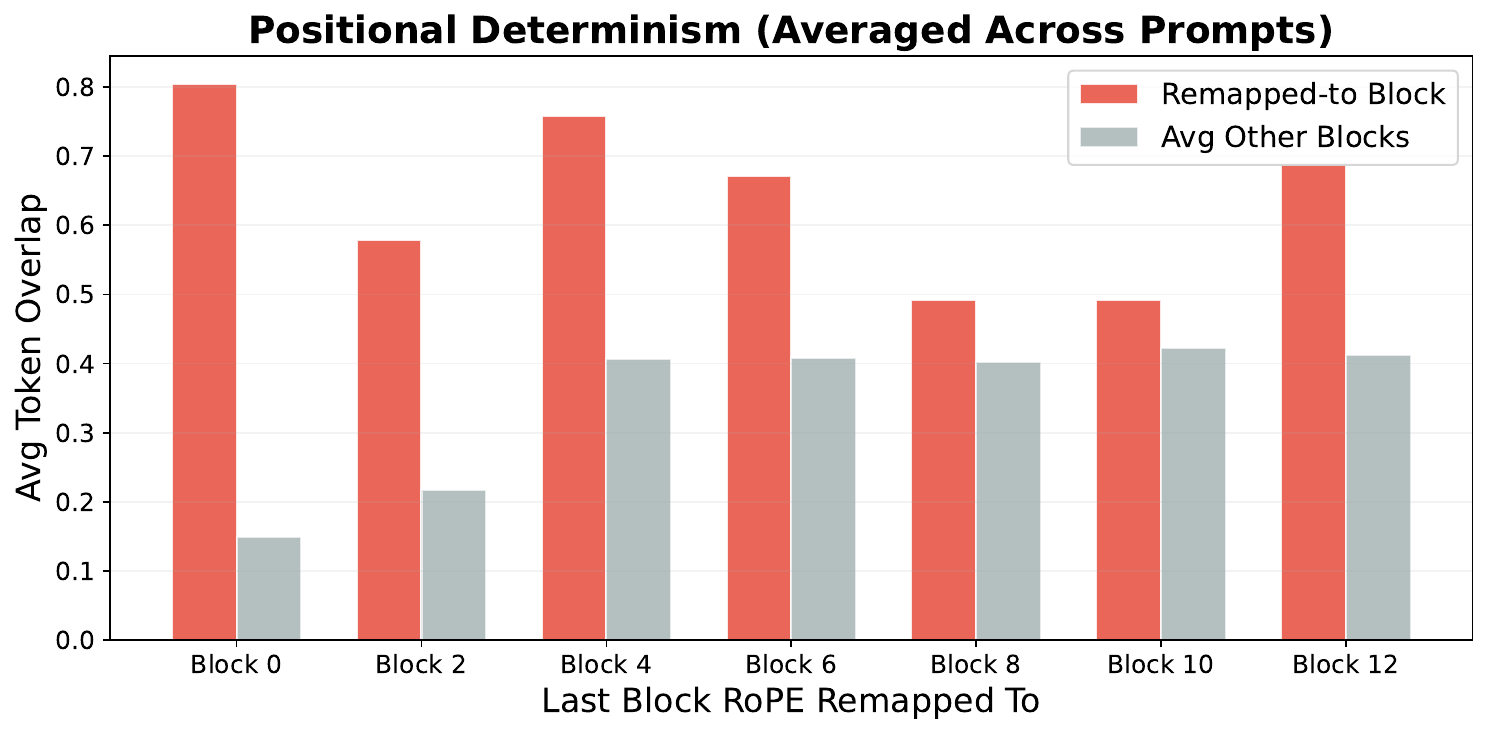}
        \caption{\textbf{RoPE Positional Control.} Generation behavior under RoPE index remapping. The generated last block moves toward the content associated with the RoPE coordinates it receives, showing that position itself shapes generation.}
        \label{fig:positional_observation}
    \end{minipage}
    \hfill
    \begin{minipage}[t]{0.48\linewidth}
        \centering
        \includegraphics[width=\linewidth]{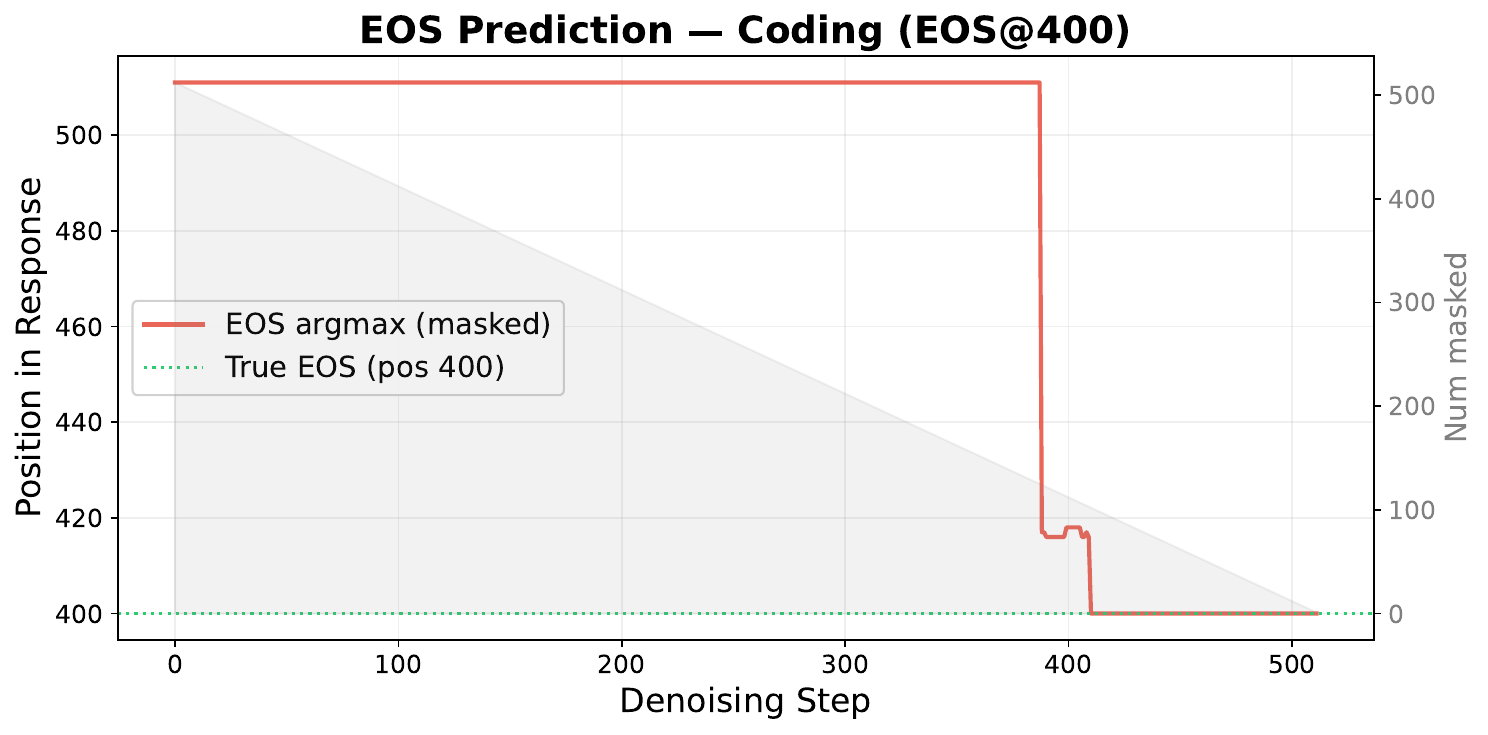}
        \caption{\textbf{Terminal Position Signal.} EOS prediction during denoising. The red curve marks the \texttt{[MASK]} position with the highest EOS probability, and the green dotted line marks the final decoded EOS position. The early peak near the end indicates that the final RoPE position provides a terminal signal.}
        \label{fig:eos_observation}
    \end{minipage}
\end{figure}

\subsection{\texttt{[MASK]} tokens contain structural information}
\label{sec:mask_structural_info}
\label{sec:rope_observation}
\label{sec:eos_observation}

\noindent\textbf{\texttt{[MASK]} tokens carry RoPE positional information.}\quad
Despite similar feature representations, different \texttt{[MASK]} positions still exhibit distinct attention behavior in Fig.~\ref{fig:mask_attention}. This suggests that their role is not determined only by the shared \texttt{[MASK]} embedding, since attention computation also combines token states with RoPE positional encoding. Therefore, \textbf{it might not be the feature representation of \texttt{[MASK]} tokens that matter, but the positional information they carry.} To verify this assumption, we test whether a generated block is determined mainly by its token state and history, or by the positional information carried by \texttt{[MASK]} tokens. As seen in Fig.~\ref{fig:positional_observation}, when decoding the last \texttt{[MASK]} block, we replace its positional information with that of the \(i\)-th previous block. The decoded content is significantly more similar to the content generated at the \(i\)-th previous substituted position, indicating that positional information strongly guides the semantics of \texttt{[MASK]} token decoding.

\noindent\textbf{The last position indicates response terminal.}\quad
As discussed above, we find that the positional information carried by \texttt{[MASK]} tokens plays an essential role in decoding. Among all positions, the final RoPE position, corresponding to the maximum sequence length configured for decoding, is particularly important because it provides a terminal signal for predicting the end-of-sequence (EOS) token. To verify this, we visualize the position with the highest EOS probability under greedy decoding in Fig.~\ref{fig:eos_observation}. In the early stages of denoising, this position is strongly correlated with the final position, i.e., position 512, suggesting that the final RoPE position encodes the terminal signal for response generation.
This observation also explains the repeated-final-fragment phenomenon discussed in Sec.~\ref{sec:intro}. Without access to the final positional signal, block dLLMs may fail to accurately infer the EOS location. Consequently, the model may not recognize when to terminate generation and may repeatedly generate the final fragment.

\section{Method}
\label{sec:method}

\begin{figure}[t]
    \centering
    \includegraphics[width=\linewidth]{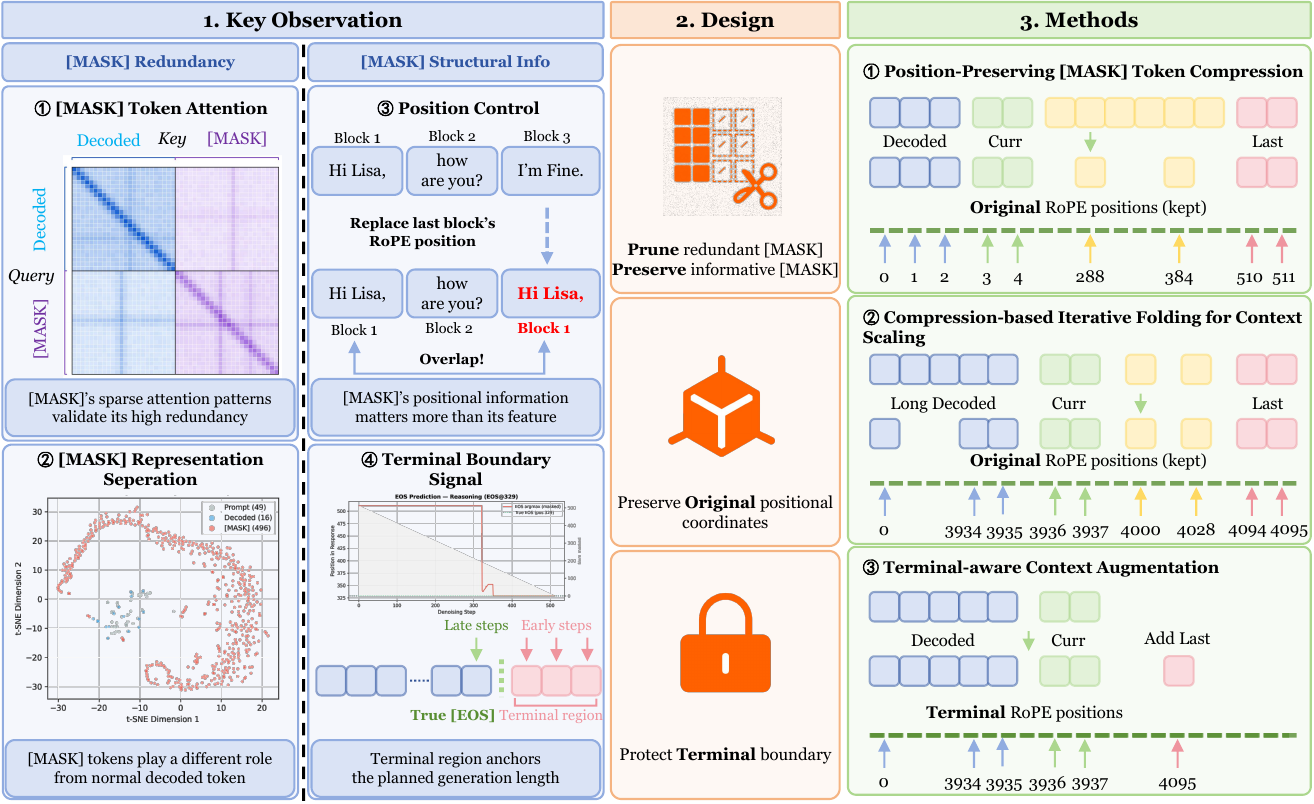}
    \caption{Overview of Elastic-dLLM. The observations show that \texttt{[MASK]} tokens contain both redundancy and structural information. Elastic-dLLM turns this finding into three training-free mechanisms: position-preserving \texttt{[MASK]} token compression, iterative folding for context scaling, and terminal-aware context augmentation for block dLLMs.}
    \label{fig:overview}
\end{figure}

\paragraph{Overview.}
Elastic-dLLM is designed from the analysis in Section~\ref{sec:observation}, with the overall framework illustrated in Fig.~\ref{fig:overview}. The attention and representation analyses show that \texttt{[MASK]} tokens introduce substantial redundant computation, while the RoPE and EOS analyses show that these tokens also carry structural information. Therefore, Elastic-dLLM follows a simple principle: compact redundant \texttt{[MASK]} computation while preserving the positional and terminal signals that guide decoding.

As summarized in Fig.~\ref{fig:overview}, Elastic-dLLM contains three training-free mechanisms. First, it performs position-preserving \texttt{[MASK]} token compression for full-sequence dLLMs, reducing the active context while keeping retained tokens at their original RoPE coordinates. Second, it extends the same compression view to iterative folding, where over-length decoded history is compressed together with \texttt{[MASK]} tokens for context scaling. Third, it performs terminal-aware context augmentation for block dLLMs by adding one protected non-decoding \texttt{[MASK]} token at the final RoPE position.

\subsection{Position-Preserving \texttt{[MASK]} Token Compression}
\label{sec:mask_token_pruning}

In a full-sequence dLLM, each denoising step processes the prompt and the preallocated generation context together. We divide the generation context into fixed-size blocks. At a given denoising stage, the input contains decoded history, the current block being denoised, remaining \texttt{[MASK]} blocks, and the terminal block. Elastic-dLLM compresses this context as follows:
\[
\begin{array}{ll}
\textbf{Original input:} & \text{Decoded history} \mid \text{Current block} \mid \text{[MASK] blocks} \mid \text{Terminal block}, \\
\textbf{Elastic input:} & \text{Decoded history} \mid \text{Current block} \mid \text{Sampled [MASK] tokens} \mid \text{Terminal block}.
\end{array}
\]

More formally, let the prompt positions be $\mathcal{P}$ and let the generation context be partitioned into blocks $\mathcal{B}_1,\ldots,\mathcal{B}_M$. At the current stage, $\mathcal{B}_c$ is the block being denoised, $\mathcal{H}_{<c}=\mathcal{B}_1,\ldots,\mathcal{B}_{c-1}$ denotes decoded history, and $\mathcal{B}_M$ is the terminal block. For a retained \texttt{[MASK]} block $\mathcal{B}_j$, let $\operatorname{Uniform}_r(\mathcal{B}_j)$ return $r$ uniformly sampled positions from that block. Elastic-dLLM keeps:
\[
    \mathcal{S}_{\mathrm{target}}
    =
    \mathcal{H}_{<c}
    \cup
    \mathcal{B}_c
    \cup
    \mathcal{B}_M
    \cup
    \bigcup_{j \in \{c+1,\ldots,M-1\}}
    \operatorname{Uniform}_r(\mathcal{B}_j),
\]
and the actual model input uses:
\[
    \mathcal{S}
    =
    \mathcal{P}
    \cup
    \mathcal{S}_{\mathrm{target}}.
\]
The set union makes the rule well-defined even when the current block is close to the end of generation. 

This retention rule follows directly from the observations in Section~\ref{sec:observation}. The decoded history and current block are left dense to preserve local semantic continuity and the tokens being updated at the current stage. The terminal block is also protected, since Fig.~\ref{fig:eos_observation} shows that the final RoPE region provides a terminal signal for EOS prediction. By contrast, middle future \texttt{[MASK]} blocks exhibit repeated representation patterns, making them suitable for compression; retaining a sparse trace is sufficient to preserve coarse positional coverage.

The RoPE observation in Fig.~\ref{fig:positional_observation} further determines how the compressed layout should be represented. A shortened input alone is not enough: if retained tokens are re-indexed to consecutive positions, the model receives a different positional prior. Elastic-dLLM therefore reduces token count while preserving the positional skeleton of the original generation context. Every retained token uses its original RoPE index. Let $h_i$ be the hidden state of a retained token with original position $i$, and let $\operatorname{rank}_{\mathcal{S}}(i)$ be the compact order of this token inside the compressed context. Elastic-dLLM applies RoPE as
\[
    Q_i = R(i) W_Q h_i,
    \qquad
    K_i = R(i) W_K h_i,
\]
instead of the compressed-coordinate variant
\[
    Q_i = R(\operatorname{rank}_{\mathcal{S}}(i)) W_Q h_i,
    \qquad
    K_i = R(\operatorname{rank}_{\mathcal{S}}(i)) W_K h_i.
\]
Here $R(\cdot)$ denotes the RoPE rotation at a given position.

\subsection{Compression-based Iterative Folding for Context Scaling}
\label{sec:test_time_scaling}

Position-preserving \texttt{[MASK]} token compression reduces the active context while keeping decoded history dense. For context scaling, however, decoded history also grows as generation proceeds and can eventually exceed the maximum input length or the active context budget. Elastic-dLLM therefore applies iterative folding, shown as the second mechanism in Fig.~\ref{fig:overview}. At each denoising stage, it folds old decoded blocks while applying the same future \texttt{[MASK]} compression rule from Section~\ref{sec:mask_token_pruning}:
\[
\begin{array}{ll}
\textbf{Original input:} &
\text{Decoded history} \mid \text{Current block} \mid \text{[MASK] blocks} \mid \text{Terminal block}, \\
\textbf{Elastic input:} &
\text{Folded history} \mid \text{Current block} \mid \text{Sampled [MASK] tokens} \mid \text{Terminal block}.
\end{array}
\]
All retained tokens, whether decoded or still masked, keep their original RoPE coordinates. The model therefore processes a compact active context whose positional skeleton still spans the longer generation context.

Let $\mathcal{H}_{<c}=\mathcal{B}_1,\ldots,\mathcal{B}_{c-1}$ denote the decoded blocks before the current block $\mathcal{B}_c$. If the decoded history fits the active budget, it is kept dense. Once folding is needed, Elastic-dLLM preserves the most recent $K$ decoded blocks in full and compresses older decoded blocks into sparse representatives:
\[
    \operatorname{Fold}(\mathcal{H}_{<c})
    =
    \bigcup_{j=c-K}^{c-1} \mathcal{B}_j
    \cup
    \bigcup_{j < c-K} \operatorname{Select}_{f}(\mathcal{B}_j).
\]
Here $K$ is the number of recent dense blocks. For each older decoded block $\mathcal{B}_j$, $\operatorname{Select}_{f}(\mathcal{B}_j)$ keeps $f$ representative tokens. It always preserves the first and last token of the block and uses uniformly spaced samples for any remaining slots:
\[
    \operatorname{Select}_{f}(\mathcal{B}_j)
    =
    \operatorname{Endpoints}(\mathcal{B}_j)
    \cup
    \operatorname{Uniform}_{f-2}(\mathcal{B}_j).
\]
This simple rule keeps a coarse trace of the decoded history, preserves block-level continuity, and avoids introducing an additional learned scorer or confidence heuristic.

The resulting active set for context scaling is
\[
    \mathcal{S}^{\mathrm{scale}}_c
    =
    \operatorname{Fold}(\mathcal{H}_{<c})
    \cup
    \mathcal{B}_c
    \cup
    \mathcal{B}_M
    \cup
    \bigcup_{j \in \{c+1,\ldots,M-1\}}
    \operatorname{Uniform}_r(\mathcal{B}_j),
\]
where $\mathcal{B}_c$ is the current denoising block and is kept dense, $\mathcal{B}_M$ is the terminal block and is also kept dense as the EOS boundary signal, and $\operatorname{Uniform}_r(\mathcal{B}_j)$ follows the position-preserving \texttt{[MASK]} sampling rule in Section~\ref{sec:mask_token_pruning} for each middle future block. After $\mathcal{B}_c$ is denoised and committed, the model advances to $\mathcal{B}_{c+1}$ and rebuilds the folded view. This gives Elastic-dLLM a compact active context while preserving three signals needed for long generation: local continuity from recent dense history, coarse coverage of older decoded blocks, and the terminal signal from the final block.

\subsection{Terminal-aware Context Augmentation}
\label{sec:block_diffusion_adaptation}

Block dLLMs~\citep{llada2} reduce the cost of full-sequence denoising by generating one block at a time. They are still different from autoregressive models because decoding is performed under a planned generation length. However, the active input at each step contains only the decoded history and the current block, so the model does not observe the final \texttt{[MASK]} position that indicates where the response should end. The EOS analysis in Fig.~\ref{fig:eos_observation} suggests that this terminal signal is useful during denoising. We therefore introduce a minimal context augmentation that restores this signal without reintroducing a long \texttt{[MASK]} context, as summarized in Fig.~\ref{fig:overview}.

At each block-decoding step, Elastic-dLLM appends one protected \texttt{[MASK]} token after the current block:
\[
\begin{array}{ll}
\textbf{Block dLLM:} & \text{Decoded} \mid \text{Current block}, \\
\textbf{Elastic block dLLM:} & \text{Decoded} \mid \text{Current block} \mid \text{Protected terminal token}.
\end{array}
\]
Let $\mathcal{H}_{<t}$ denote the prompt and previously generated history, let $\mathcal{G}_t$ denote the current block, and let $\mathcal{A}_t=\{a_t\}$ denote the protected terminal token. The active input becomes
\[
    \mathcal{S}^{\mathrm{aug}}_t
    =
    \mathcal{H}_{<t}
    \cup
    \mathcal{G}_t
    \cup
    \mathcal{A}_t.
\]
The protected token participates in attention but is not decoded. It is also not assigned the next local position. Instead, under planned generation length $N_{\mathrm{gen}}$, it receives the final RoPE coordinate
\[
    \pi(a_t)
    =
    N_{\mathrm{gen}}.
\]
Thus, $a_t$ is adjacent to the current block in the compact input, but positionally it points to the end of the planned generation context. After $\mathcal{G}_t$ is committed, the anchor is discarded and rebuilt for the next block. Since this augmentation adds only one token to each active input, its computational overhead is negligible, yet it provides a persistent terminal signal for block dLLMs.

\section{Experiments}
\label{sec:experiments}

\subsection{Experimental Setting}
\label{sec:setup}

\begin{table*}[t]
    \centering
    \caption{Accuracy (\%) of full-sequence dLLMs on MATH500~\citep{math500}, GSM8K~\citep{gsm8k}, ASDiv~\citep{asdiv}, Countdown~\citep{countdown}, HumanEval~\citep{humaneval}, and IFEval~\citep{ifeval}.}
    \label{tab:full_seq_main}
    \small
    \setlength{\tabcolsep}{4pt}
    \renewcommand{\arraystretch}{0.9}
    \begin{tabular}{l cccc c c}
    \toprule
    & \multicolumn{4}{c}{Mathematical Reasoning} & \multicolumn{1}{c}{Code} & \multicolumn{1}{c}{Instruction} \\
    \cmidrule(lr){2-5} \cmidrule(lr){6-6} \cmidrule(lr){7-7}
    & MATH500 & GSM8K & ASDiv & Countdown & HumanEval & IFEval \\
    \midrule
    \multicolumn{7}{l}{\textbf{\textit{LLaDA-8B-Instruct}}} \\
    \midrule
    Baseline~\citep{llada} & 30.20 & 81.88 & 80.95 & 26.56 & 40.24 & 52.68 \\
    DPad~\citep{dpad} & 29.40 & 81.35 & 80.65 & 25.39 & 38.41 & 51.20 \\
    SparseD~\citep{sparsed} & 29.80 & 81.50 & 80.52 & 25.78 & 39.02 & 52.31 \\
    Elastic  & 30.20 & 80.74 & 80.87 & 38.67 & 41.46 & 52.31 \\
    \midrule
    \multicolumn{7}{l}{\textbf{\textit{LLaDA-1.5}}} \\
    \midrule
    Baseline~\citep{llada1.5} & 32.20 & 80.36 & 81.34 & 25.39 & 41.46 & 58.41 \\
    DPad~\citep{dpad} & 31.00 & 79.98 & 80.95 & 23.83 & 39.63 & 57.86 \\
    SparseD~\citep{sparsed} & 31.80 & 80.06 & 81.17 & 24.61 & 40.24 & 57.67 \\
    Elastic  & 32.20 & 79.76 & 81.17 & 29.69 & 41.46 & 58.04 \\
    \bottomrule
    \end{tabular}
\end{table*}

\paragraph{Models} We evaluate Elastic-dLLM on both full-sequence and block dLLMs. For full-sequence dLLMs, we use LLaDA-8B-Instruct~\citep{llada} and LLaDA-1.5~\citep{llada1.5}. For block dLLMs, we use LLaDA2.0-mini~\citep{llada2}. Each model is compared under its original decoding procedure and the Elastic configuration described in Section~\ref{sec:method}.

\paragraph{Baselines} For full-sequence dLLMs, we compare with the original decoding baseline, DPad~\citep{dpad}, and SparseD~\citep{sparsed}. DPad reduces redundant suffix computation while SparseD applies sparse attention patterns for dLLMs. For compatibility evaluation, we combine Elastic-dLLM with Fast-dLLM~\citep{fastdllm}, which introduces approximate \textbf{KV cache reuse} and \textbf{confidence-aware parallel decoding}.

\paragraph{Datasets} We evaluate mathematical reasoning on MATH500~\citep{math500}, GSM8K~\citep{gsm8k}, ASDiv~\citep{asdiv}, and Countdown~\citep{countdown}; code generation on HumanEval~\citep{humaneval}; instruction following on IFEval~\citep{ifeval}; and long-form generation on LongWriter-Bench~\citep{longwriter}. For Countdown, we follow the evaluation prompt used by d1~\citep{d1}. All evaluations use 0-shot prompting. 

\paragraph{Implementation Details} For MATH500~\citep{math500}, GSM8K~\citep{gsm8k}, ASDiv~\citep{asdiv}, Countdown~\citep{countdown}, HumanEval~\citep{humaneval}, and IFEval~\citep{ifeval}, LLaDA-8B-Instruct~\citep{llada} and LLaDA-1.5~\citep{llada1.5} use 512 denoising steps, block size 32, maximum generation length 512, and temperature 0.0. LLaDA2.0-mini~\citep{llada2} uses block size 32, maximum generation length 1024, temperature 0.0, EOS early stopping, and one protected terminal token ($|\mathcal{A}_t|=1$). All experiments run on NVIDIA RTX 3090 GPUs.

\subsection{Main Results on Full-Sequence dLLMs}
\label{sec:main_results_fullseq}

\paragraph{Standard benchmark results.}
Table~\ref{tab:full_seq_main} evaluates context compression on LLaDA-8B-Instruct~\citep{llada} and LLaDA-1.5~\citep{llada1.5}. Elastic-dLLM keeps the current and terminal regions dense, samples redundant \texttt{[MASK]} tokens, and preserves original RoPE coordinates.

Elastic-dLLM remains close to the original baseline on tested benchmarks, while improving Countdown~\citep{countdown}. Compared with DPad~\citep{dpad} and SparseD~\citep{sparsed}, it gives a stronger compact-context accuracy profile, especially on Countdown and HumanEval.

Table~\ref{tab:elastic_fast} shows that this layout-level compression is compatible with Fast-dLLM~\citep{fastdllm}. On LLaDA-1.5~\citep{llada1.5}, Elastic-dLLM gives a $1.52\times$ speedup, and Elastic+Fast-dLLM reaches $7.28\times$ with similar Countdown~\citep{countdown} and IFEval~\citep{ifeval} scores.

\paragraph{Test-time context scaling.}
Table~\ref{tab:longwriter_context_scaling} evaluates iterative folding on LongWriter-Bench~\citep{longwriter}. The key stress range is $[4k,20k)$, where decoded history exceeds the normal active context budget and must be folded into a compact positional trace. Elastic-dLLM is consistently better than the base layout across LongWriter-Bench ranges, showing that the compact context is effective for long-form generation. More importantly, the $S_l/S_q$ scores in the $[4k,20k)$ range show that folding remains effective where the baseline full-sequence layout struggles, supporting the second mechanism: Elastic-dLLM scales long-form generation by compressing decoded history after it exceeds the available context budget.

\begin{table}[t]
    \centering
    \begin{minipage}[t]{0.48\linewidth}
        \centering
        \caption{Elastic-dLLM with Fast-dLLM~\citep{fastdllm} on LLaDA-1.5~\citep{llada1.5}.}
        \label{tab:elastic_fast}
        \small
        \renewcommand{\arraystretch}{0.86}
        \setlength{\tabcolsep}{3pt}
        \begin{tabular}{lccc}
        \toprule
        Method & Countdown & IFEval & Speedup \\
        \midrule
        Baseline & 25.39 & 58.41 & 1.00$\times$ \\
        Elastic & 29.69 & 58.04 & 1.52$\times$ \\
        Elastic + Fast-dLLM & 29.30 & 59.70 & 7.28$\times$ \\
        \bottomrule
        \end{tabular}
    \end{minipage}
    \hfill
    \begin{minipage}[t]{0.48\linewidth}
        \centering
        \caption{LongWriter-Bench~\citep{longwriter} results for iterative folding.}
        \label{tab:longwriter_context_scaling}
        \small
        \renewcommand{\arraystretch}{0.84}
        \setlength{\tabcolsep}{1.5pt}
        \resizebox{\linewidth}{!}{
        \begin{tabular}{l|cc|cc|cc|cc}
        \toprule
          & \multicolumn{2}{c|}{\textbf{[0, 500)}} & \multicolumn{2}{c|}{\textbf{[500, 2k)}} & \multicolumn{2}{c|}{\textbf{[2k, 4k)}} & \multicolumn{2}{c}{\textbf{[4k, 20k)}} \\
         \cmidrule(lr){2-3} \cmidrule(lr){4-5} \cmidrule(lr){6-7} \cmidrule(lr){8-9}
          & $S_l$ & $S_q$ & $S_l$ & $S_q$ & $S_l$ & $S_q$ & $S_l$ & $S_q$ \\
        \midrule
        8B Base & 84.0 & 48.5 & 75.7 & 40.6 & 51.2 & 18.6 & 23.4 & 4.2 \\
        8B Elastic & 86.0 & 62.3 & 91.2 & 57.3 & 87.9 & 42.1 & 55.6 & 22.8 \\
        1.5 Base & 76.3 & 58.6 & 85.8 & 59.9 & 50.7 & 34.6 & 28.7 & 5.9 \\
        1.5 Elastic & 94.9 & 71.3 & 92.5 & 65.6 & 84.7 & 46.2 & 77.4 & 39.3 \\
        \bottomrule
        \end{tabular}
        }
    \end{minipage}
\end{table}

\subsection{Main Results on Block dLLMs}
\label{sec:main_results_blockdiff}

Table~\ref{tab:block_diffusion_main} reports the main results for block dLLMs on LLaDA2.0-mini~\citep{llada2}. Elastic-dLLM adds only one protected terminal \texttt{[MASK]} token at the final RoPE position, which gives the current block access to the planned generation boundary with negligible context overhead.
The single-token terminal signal improves MATH500~\citep{math500}, ASDiv~\citep{asdiv}, Countdown~\citep{countdown}, and IFEval~\citep{ifeval}, with the largest gain on Countdown~\citep{countdown}. Since the method adds only one token, these gains indicate that the terminal position carries useful generation information for block dLLMs.

\begin{table*}[t]
    \centering
    \caption{Accuracy (\%) of LLaDA2.0-mini~\citep{llada2}. Elastic appends a single protected \texttt{[MASK]} token at the terminal RoPE position. $\Delta$ denotes the change from the baseline.}
    \label{tab:block_diffusion_main}
    \small
    \begin{tabular}{l cccc c c}
    \toprule
    & \multicolumn{4}{c}{Mathematical Reasoning} & \multicolumn{1}{c}{Code} & \multicolumn{1}{c}{Instruction} \\
    \cmidrule(lr){2-5} \cmidrule(lr){6-6} \cmidrule(lr){7-7}
    & MATH500 & GSM8K & ASDiv & Countdown & HumanEval & IFEval \\
    \midrule
    Baseline & 47.20 & 90.67 & 86.20 & 41.41 & 77.44 & 83.55 \\
    Elastic  & 49.60 & 90.67 & 86.72 & 65.62 & 77.44 & 85.03 \\
    \midrule
    $\Delta$ & +2.40 & 0.00 & +0.52 & +24.21 & 0.00 & $+$1.48 \\
    \bottomrule
    \end{tabular}
    \vspace{-10pt}
\end{table*}

\subsection{Ablation Studies}
\label{sec:ablation}

\paragraph{RoPE coordinate preservation.}
Table~\ref{tab:rope_ablation} compares two RoPE assignments after context compression on LLaDA-8B-Instruct~\citep{llada} and LLaDA-1.5~\citep{llada1.5}. Compact RoPE reassigns retained tokens to consecutive indices in the compressed context, while Elastic-dLLM keeps their original RoPE coordinates from the full generation context.

\begin{wraptable}{r}{6.4cm}
    \vspace{-5mm}
    \centering
    \caption{RoPE ablation under the same compressed token layout.}
    \label{tab:rope_ablation}
    \small
    \renewcommand{\arraystretch}{0.84}
    \setlength{\tabcolsep}{2pt}
    \begin{tabular}{lcc}
    \toprule
    Method & Countdown & IFEval \\
    \midrule
    \multicolumn{3}{c}{LLaDA-8B-Instruct} \\
    \midrule
    Baseline & 26.56 & 52.68 \\
    Compact RoPE & 17.97 & 43.81 \\
    Elastic & 38.67 & 52.31 \\
    \midrule
    \multicolumn{3}{c}{LLaDA-1.5} \\
    \midrule
    Baseline & 25.39 & 58.41 \\
    Compact RoPE & 17.58 & 48.61 \\
    Elastic & 29.69 & 58.04 \\
    \bottomrule
    \end{tabular}
    \vspace{-4mm}
\end{wraptable}

Compact RoPE performs much worse than Elastic-dLLM on Countdown~\citep{countdown} and IFEval~\citep{ifeval}, even though it uses the same compressed token layout. This gap shows that compression alone is insufficient. The retained tokens must preserve their original RoPE coordinates rather than adopt the new indices induced by the compact context, which matches the positional-control observation in Section~\ref{sec:rope_observation}.

\paragraph{Terminal anchor design.}
Table~\ref{tab:ablation} ablates two terminal-anchor choices in LLaDA2.0-mini~\citep{llada2}: attention pattern and anchor content. The default setting uses \texttt{[MASK]} as the anchor content and main-only-sees attention (mos), where the main sequence can attend to the anchor but the anchor cannot attend back to the main sequence. The main sequence uses the terminal RoPE position as a clean boundary signal while preventing the anchor from being updated by the incomplete denoising state.

\begin{wraptable}{r}{6.4cm}
    \vspace{-5mm}
    \centering
    \caption{Ablation of terminal anchor design choices on LLaDA2.0-mini~\citep{llada2}.}
    \label{tab:ablation}
    \scriptsize
    \setlength{\tabcolsep}{2pt}
    \renewcommand{\arraystretch}{0.84}
    \resizebox{\linewidth}{!}{
    \begin{tabular}{lcccc}
    \toprule
    Config & Content & Attn & Countdown & MATH500 \\
    \midrule
    Baseline        & --- & --- & 41.41 & 47.20 \\
    \midrule
    \multicolumn{5}{l}{\emph{Attention pattern}} \\
    Elastic (default)  & \texttt{[M]} & mos & \textbf{65.62} & \textbf{49.60} \\
    Bidirectional   & \texttt{[M]} & bidir & 2.34 & 37.60 \\
    \midrule
    \multicolumn{5}{l}{\emph{Anchor content}} \\
    \texttt{[MASK]} (default) & \texttt{[M]} & mos & \textbf{65.62} & \textbf{49.60} \\
    \texttt{[EOS]} token      & \texttt{[EOS]} & mos & 0.00 & 2.40 \\
    \bottomrule
    \end{tabular}
    }
    \vspace{-4mm}
\end{wraptable}

The results show that the anchor is useful only when it remains a protected boundary signal. Changing mos to bidirectional attention causes a severe drop, especially on Countdown~\citep{countdown}. In this setting, the anchor also attends to the main sequence and becomes entangled with the current denoising state, so it no longer serves as a clean terminal reference. Replacing \texttt{[MASK]} with \texttt{[EOS]} is also harmful, since \texttt{[EOS]} is a strong termination token rather than a neutral non-decoding placeholder. \textbf{The terminal anchor should be a protected non-decoding \texttt{[MASK]} token with one-way visibility from the main sequence.}

\section{Conclusion}
\label{sec:conclusion}

We presented Elastic-dLLM, a training-free approach for making diffusion LLM inference more efficient by rethinking the role of \texttt{[MASK]} tokens. Our analysis shows that \texttt{[MASK]} tokens introduce substantial redundant computation, but also carry structural signals through their RoPE positions and the terminal generation boundary. Based on this observation, Elastic-dLLM compresses the active context while preserving original positions, extends the same idea to iterative folding for long generation, and restores terminal awareness in block dLLMs with a single protected \texttt{[MASK]} token.


\bibliographystyle{plainnat}
\bibliography{ghost}



\end{document}